# CSSegNet: Fine-Grained Cardiac Structures Segmentation Using Dilated Pyramid Pooling in U-net

Fei Feng [1] and Jiajia Luo [1] (✉)

[1]University of Michigan – Shanghai Jiao Tong University Joint Institute, Shanghai Jiao Tong University Shanghai, 200240, China
jiajia.luo@sjtu.edu.cn

**Abstract.** Cardiac structure segmentation plays an important role in medical analysis procedures. Images' blurred boundaries issue always limits the segmentation performance. To address this difficult problem, we presented a novel network structure which embedded dilated pyramid pooling block in the skip connections between network's encoding and decoding stage. A dilated pyramid pooling block is made up of convolutions and pooling operations with different vision scopes. Equipped the model with such module, it could be endowed with multi-scales vision ability. Together combining with other techniques, it included a multi-scales initial features extraction and a multi-resolutions' prediction aggregation module. As for backbone feature extraction network, we referred to the basic idea of Xception network which benefited from separable convolutions. Evaluated on the Post 2017 MICCAI-ACDC challenge phase data, our proposed model could achieve state-of-the-art performance in left ventricle (LVC) cavities and right ventricle cavities (RVC) segmentation tasks. Results revealed that our method has advantages on both geometrical (Dice coefficient, Hausdorff distance) and clinical evaluation (Ejection Fraction, Volume), which represent closer boundaries and more statistically significant separately.

## 1 Introduction

Cardiac magnetic resonance imaging (CMRI) is an important technique used for cardiac pathological and morphological evaluation in different cardiac treatment stages. Because of its high contrast between different tissues, it is usually the first choice for the cardiac tissues' segmentation. Cardiac structures segmentation is a common procedure for diastolic and systolic volumes measurements, and ejection fraction measurements. The related segmentation regions are left ventricle (LVC) cavities, right ventricle cavities (RVC), and left ventricle myocardium (LVM) between them. Unlike qualitative malignant or healthy analysis, cardiac segmentation on different region boundaries is a time-consuming operation and prone to inconsistency between radiologists. Existing difficulties on the cardiac structure segmentation include: 1) Huge variation of the LVC, RVC and LVM profiles at different instant because of the cardiac movement;

2) Blurred boundary and low contrast between region of interests and background. As the growth on the demand of CMRI examination, it causes a huge challenge to limited medical resources. Therefore, computer aided techniques are highly anticipated to solve this dilemma.

Both automated or semi-automated segmentation methods have been investigated for a while. Especially, CNN based automated segmentation methods have been widely studied in recent years. Efficient data flow is crucial for the CNN based methods. Encode-decode model structures, which encodes the features' information in the feature extraction step, and decodes it for prediction, has been proven effective in many different scenarios [1, 2]. Inspired by human knowledge, good segmentation practice should be able to deal with images on different resolutions. Spatial pooling module was proposed to better localize the regions boundary with perceptions on difference size of field of view (FOV) on a whole image to obtain a better feature extraction[3]. Dilated convolution was investigated to improve CNN based segmentation performance on fuzzy boundary by enlarging the perception region[4]. Deep supervision ensembles different resolutions feature maps by applying supervision on smaller size intermediate inputs to have a better prediction[5].

U-net has been successfully applied in biomedical segmentation. An original U-net structure was shown in Fig.1, which has a symmetric model structure. Here we call the data path between the smallest tensors as a feature extraction shortcut path, while we call higher level shortcuts connections as feature decoding shortcut path. For U-net, 3*3 vision scale made it pay more attention to the local information but ignore the bigger scale context information. Therefore, it caused the low-quality boundary prediction.

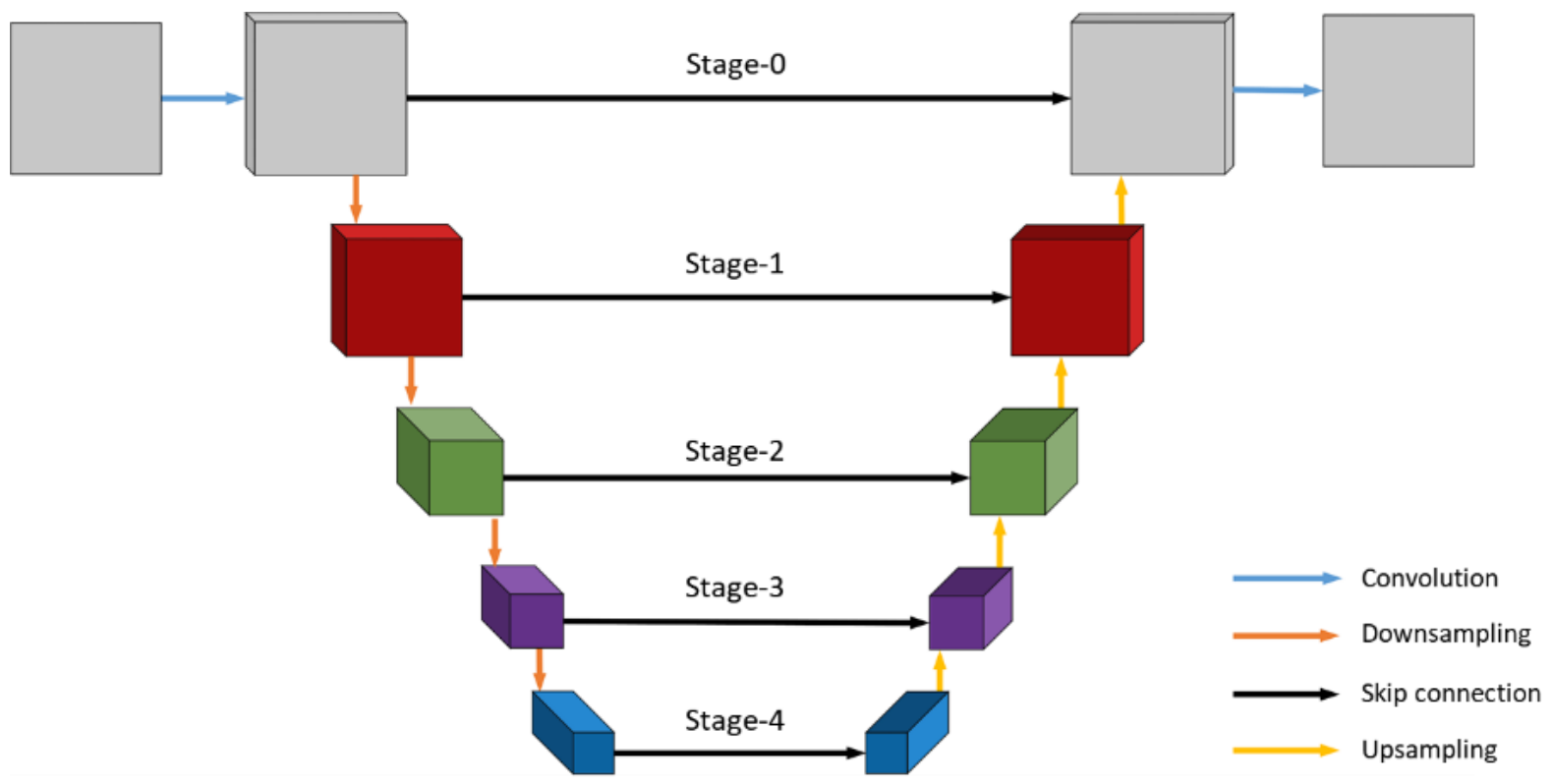

**Fig. 1.** U-net model structure.

To obtain closer boundaries of different structures from the blurred images, we proposed to insert the spatial pyramid pooling module in the shortcut dataflow on both features extraction and features decoding shortcut connections. By applying this idea into the feature map extraction stage, it could observe context information at different scales. While at prediction, it could remedy the blurred boundary issues caused by generated feature map in low resolution. Except the pooling pyramid module, other advanced ideas were also incorporated into the basic data flow. To improve segmentation performance, three common ideas need to be considered which include the original data perception, the intermediate data flow, and the final segmentation map prediction. Therefore, we also took several other actions including multi-scales concatenation and multi-resolutions supervision to improve the model's performance on the initial feature perception and at final mask prediction steps. In this paper, we designed a variant of U-net and evaluated it on the 2017 MICCAI-ACDC challenge dataset[1]. Main contributions in this work could be summarized from three aspects: 1) We proposed to embed the feature pyramid pooling module into the shortcut connections between encoder and decoder networks. By applying this module, model could be endowed the ability to view larger vision field. 2) Using initial multi-scale perception concatenation and strong supervision at final prediction stage, model's performance was improved to segment boundaries of blurred images. 3) Using only two-dimensional (2D) information, our model could achieve state-of-the-art performance on the 2017 MICCAI-ACDC challenge dataset, according to the public evaluation. We even outperformed the previous best model on some clinical important metrics.

## 2 Methods

### 2.1 Data processing

We used the data from the 2017 MICCAI-ACDC challenge. There are total 100 subjects' data for training and 50 patients' data without annotation for testing. All of them has 3D MR images taken at end diastolic (ED) and end systolic moments. There are huge variations among these data' resolution on three dimensions. As we used the 2D CNN, all the 3D data were first processed into 2D images. All images were then resampled to 1.25mm spacing in both height and width dimensions. Since these images may have different sizes, and same image has different sizes of height and width, we decided to crop the image to (256, 256) size, with centered cutting on both dimensions. Some augmentation techniques were investigated during the model training process that include affine transformation, elastic transformation, image sharpening, and contrast normalization.

### 2.2 Model structure

The conceptual structure of our proposed model is shown in Fig. 2. Our model consists of three parts which include the feature extraction (Fig. 2 left side), feature shortcut connection (Fig. 2 middle), and feature restoring (Fig. 2 right side). In the feature extraction part, we used 3 convolutions with different strides concatenation to start the initial perception. Xception[6] was used as backbone CNN structure for our model. However, we proposed to replace the identity mapping concatenation between the encoding and decoding stage with dilated spatial pooling pyramid. In the feature restoring part, stronger supervision was introduced to reduce overfitting and ensure better segmentation boundary.

[1] https://www.creatis.insa-lyon.fr/Challenge/acdc/participation.html

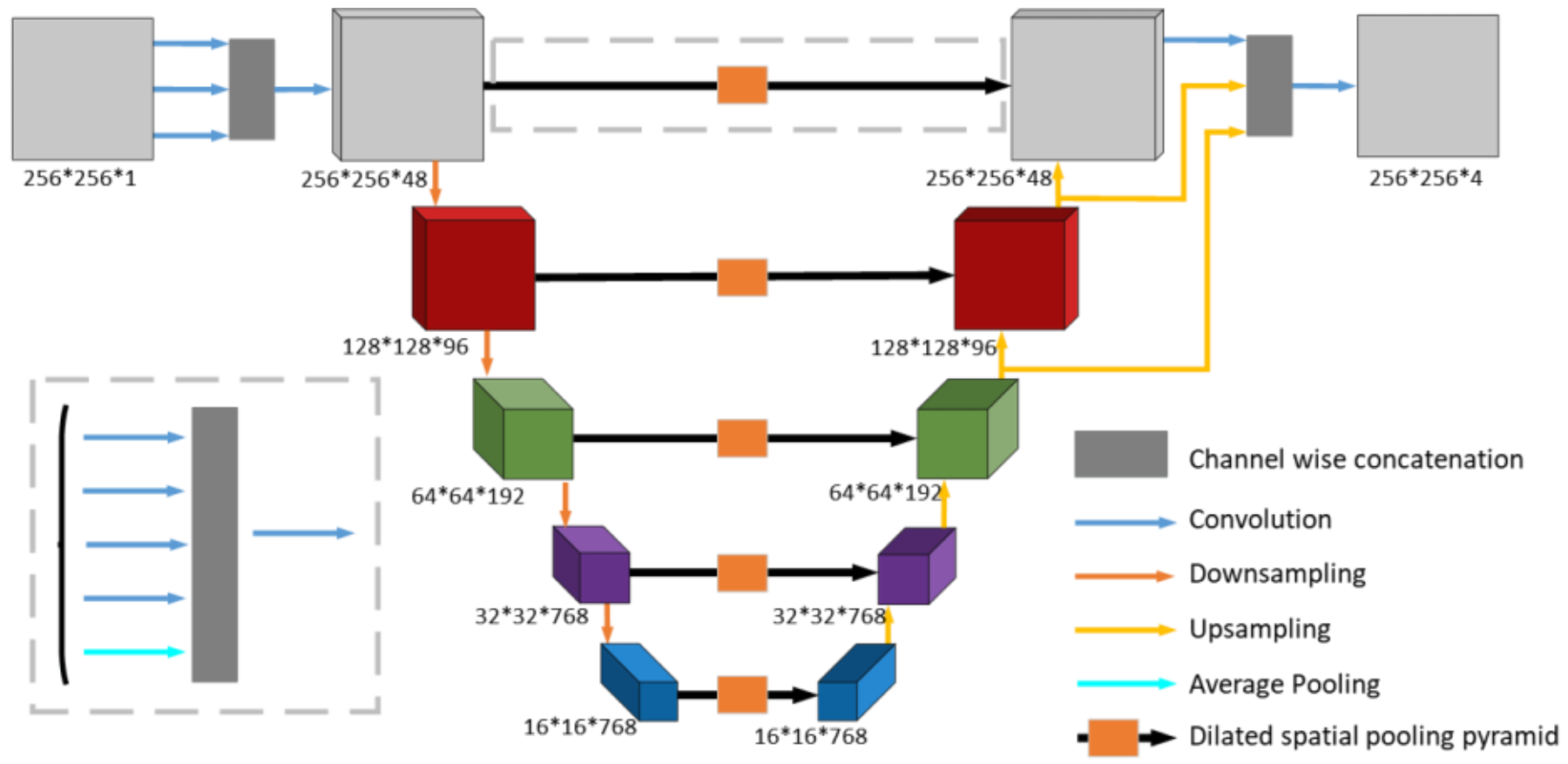

**Fig. 2.** CNN model structure we proposed to achieve more accurate cardiac structures segmentation.

**Dilated spatial pooling pyramid.** To achieve more accurate segmentation result, we adopted the idea of multi-scale views' features aggregation to make prediction. To balance the computation amount and segmentation performance, the 2D convolution kernel size was always 3*3 but sometimes with dilations. Referring to other studies[4], our pooling pyramid component each included two normal convolution, two dilated convolution and an average pooling operation, with the same number of feature channels (Table 1). But at the five different stages (Fig.2 middle), different numbers of feature channels were applied. In addition, as the feature map size was reduced by the pooling operation (Fig. 2 left), bilinear resampling was used to resample it to the original feature size. After the parallel pyramid pooling operations, feature maps were concatenated as a unitive tensor. A convolution operation was followed to integrate these feature maps and therefore to reduce the channel size. Roles of the final convolution included two aspects: 1) To mix all the feature maps extracted at different scales; and 2) To compress the feature channels to fasten the data flow. With the pyramid pooling in the shortcut connections, encoding stages' features could be fully utilized at different scales. As a result, different resolutions' details could be observed for segmentation mask prediction.

**Table 1.** Configuration summary of dilated pyramid pooling module.

| Operation | Feature size | Strides | Dilation rate |
|---|---|---|---|
| Convolution | 1*1 | 1 | No |
| Convolution | 3*3 | 1 | No |
| Convolution | 3*3 | 1 | 1 |
| Convolution | 3*3 | 1 | 2 |
| Average Pooling | 3*3 | 3 | - |

**Xception backbone and multi-resolution inputs.** Basic CNN model structure plays an important role in the whole model's performance. Based on the basic model idea presented in Xception[6], the feature extraction part was constructed as illustrated in Fig.3.

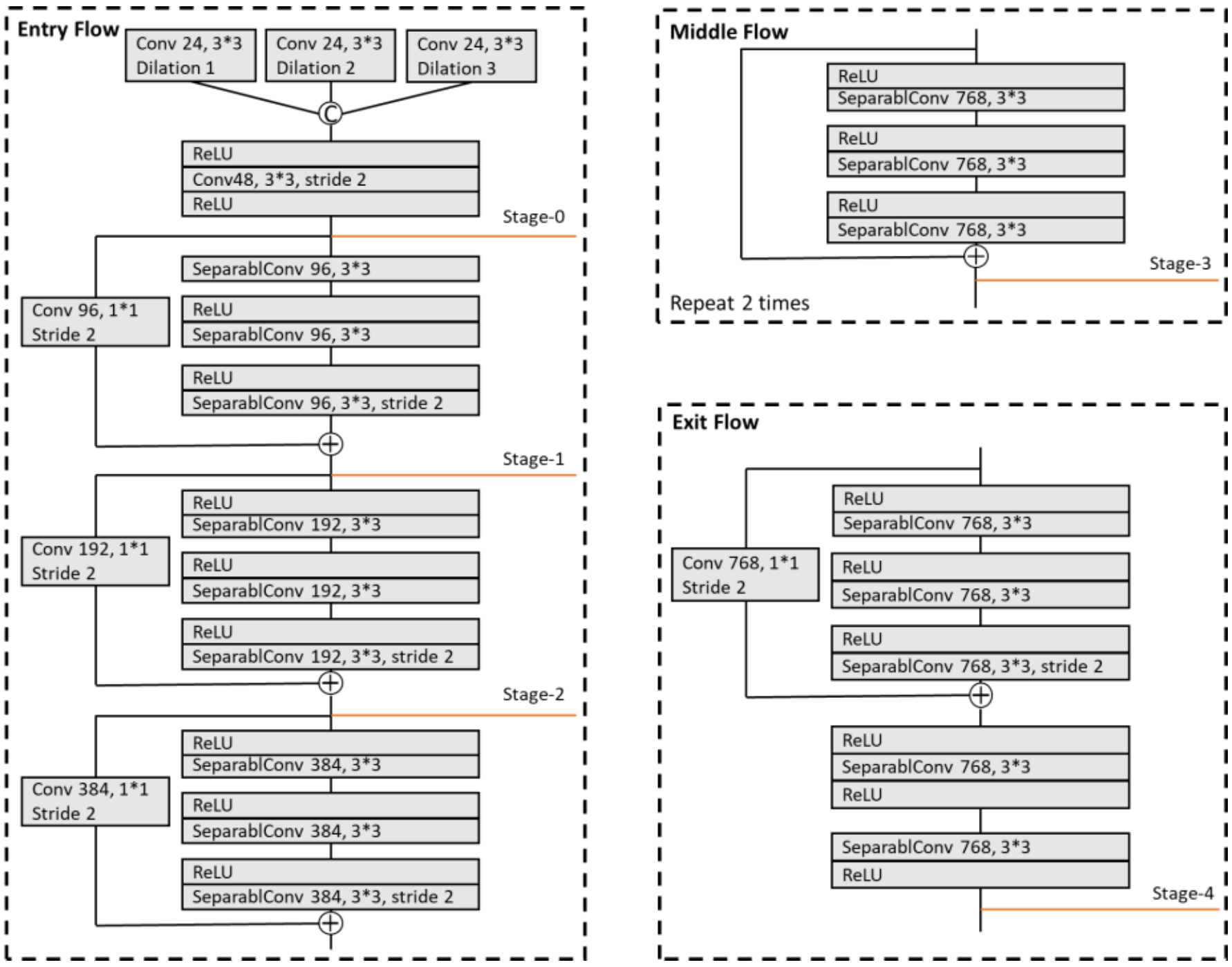

**Fig. 3.** Backbone CNN structure in feature extraction part.

Some modifications of Xception structure were made in our model. Firstly, single convolution was replaced with three scales convolution concatenation. Different with the original Xception structures, there is no image downsampling at stage-0, since we wanted to preserve more details information to fulfill better boundary segmentation. Secondly, pooling operations at stage-1 to stage-4 were changed to convolution with stride 2. Finally, to avoid overfitting, we only kept two middle flow blocks with skip connection.

**Deep supervision.** In the feature restoring part (Fig. 2 right), we ensembled the information from the top-three level feature maps to make the final prediction. We resampled the lower size feature maps as the same size as the first top level feature maps, which is the same as that of the input images and output masks. For deep feature maps, more than 4 times smaller than the initial size might be too coarse to hurt the prediction accuracy.

Generalize dice loss (GDL) was used for the loss function:

$$\text{GDL} = 1 - 2\frac{\sum_{l=1}^{N}\omega_l \sum_n r_{ln}p_{ln}}{\sum_{l=1}^{N}\omega_l \sum_n r_{ln}+p_{ln}} \quad (1)$$

N is the segmentation classes number (including background), and thus it should be 4 in our study. $r_{ln}$ and $p_{ln}$ were the ground truth and prediction annotation separately. $\omega_l$ was proposed to balance the size difference and frequency between different classes, which was constructed as $\omega_l = 1/(\sum_{l=1}^{N} r_{ln})^2$ in our model[7].

Our model was built on top of Keras (v.2.2.0) with TensorFlow (v.1.5.0) using Python (v.3.5), and it was trained with a Nvidia 1080Ti graphic card. Adam optimizer was used with default configuration except learning rate starting with 0.001.

# 3 Experimental results

To train the model, we only used the training dataset of 2017 MICCAI-ACDC challenge, which includes 100 patients' data. Using the cross-validation method, we split the dataset with 0.8:0.2 ratio for training set and validation set. Both geometrical measures and clinical measures were used to evaluate our model's performance. Geometrical measures include dice coefficient (Dice) and hausdorff distance (Hausdorff), which are consistent in all three tasks evaluation. While clinical evaluation includes volume, ejection fraction (EF), and mass measurement, which could vary in different tasks. Detailed information about metrics could be found in [8].

**Table 2.** Segmentation result on LVC structure[2].

| Methods | Dice | | Hausdorff | | EF | | Volume ED | |
|---|---|---|---|---|---|---|---|---|
| | ED | ES | ED | ES | Corr. | Bias | Corr. | Bias |
| Proposed | 0.960 | **0.920** | 6.950 | **6.950** | **0.991** | 0.540 | **0.997** | 1.840 |
| Isensee *et al.*[5] | **0.964** | **0.920** | 6.228 | 7.546 | 0.989 | 0.938 | 0.995 | 1.242 |
| Zotti *et al.*[9] | **0.964** | 0.912 | **6.180** | 8.386 | 0.990 | **-0.476** | **0.997** | 3.746 |
| Khened *et al.*[10] | 0.963 | 0.917 | 8.129 | 8.968 | 0.989 | -0.548 | **0.997** | **0.576** |
| Baumgartner *et al.*[11] | 0.963 | 0.911 | 6.526 | 9.170 | 0.988 | 0.568 | 0.995 | 1.436 |
| Wolterink *et al.*[12] | 0.961 | 0.918 | 7.515 | 9.603 | 0.988 | -0.494 | 0.993 | 3.046 |

To reduce overfitting, we saved parameters of our model which got the top five dice scores. The arithmetic mean of each result was used as the final prediction result to submit for evaluation. The top-five results for each task reported in the leaderboard were used as comparison to our model (Tables 2, 3, and 4). All results in the leaderboard should have already been accepted for publication, or their results could only be observed by participants themselves. From Table 2, we could observe our models' performance on LVC segmentation could achieved the state-of-the-art results, from both geometrical and clinical aspects, with 4 indexes achieving the best. It is worth to mention that, our predictions were made only by the 2D CNN learning while Isensee et al. combined both the 2D and 3D CNN results[5].

[2] All these results are captured in Mar 10th, 2019 on website https://acdc.creatis.insa-lyon.fr/#phase/59db86a96a3c7706f64dbfed.

**Table 3.** Segmentation result on RVC structure.

| Methods | Dice | | Hausdorff | | EF | | Volume ED | |
|---|---|---|---|---|---|---|---|---|
| | ED | ES | ED | ES | Corr. | Bias | Corr. | Bias |
| Proposed | 0.940 | 0.880 | **9.10** | 11.99 | 0.847 | -1.940 | 0.991 | 2.490 |
| Isensee *et al.*[5] | **0.947** | **0.885** | 9.39 | **11.23** | **0.894** | -2.692 | **0.993** | **0.062** |
| Zotti *et al.*[9] | 0.941 | 0.882 | 11.05 | 12.65 | 0.869 | **-0.872** | 0.986 | 2.372 |
| Zotti *et al.*[13] | 0.935 | 0.879 | 10.32 | 14.05 | 0.872 | -2.228 | 0.991 | -3.722 |
| Khened *et al.*[10] | 0.932 | 0.883 | 13.99 | 13.93 | 0.858 | -2.246 | 0.982 | -2.896 |
| Baumgartner *et al.*[11] | 0.928 | 0.872 | 12.67 | 14.69 | 0.851 | 1.218 | 0.977 | -2.290 |

**Table 4.** Segmentation result on LVM structure.

| Methods | Dice | | Hausdorff | | EF | | Volume ED | |
|---|---|---|---|---|---|---|---|---|
| | ED | ES | ED | ES | Corr. | Bias | Corr. | Bias |
| Proposed | 0.880 | 0.900 | 8.78 | **8.52** | **0.991** | -2.990 | 0.988 | -3.080 |
| Isensee *et al.*[5] | **0.897** | **0.918** | **8.56** | 8.74 | 0.989 | -1.836 | 0.984 | **-0.830** |
| Zotti *et al.*[9] | 0.886 | 0.902 | 9.59 | 9.29 | 0.980 | **1.160** | 0.986 | -1.827 |
| Khened *et al.*[10] | 0.889 | 0.898 | 9.84 | 12.58 | 0.979 | -2.572 | **0.990** | -2.873 |
| Jain *et al.*[14] | 0.882 | 0.897 | 9.75 | 11.26 | 0.986 | -4.464 | 0.989 | -11.586 |
| Baumgartner *et al.*[11] | 0.892 | 0.901 | 8.70 | 10.64 | 0.983 | -9.602 | 0.982 | -6.861 |

Segmentation performance on RVC and LVM (Tables 3 and 4) also indicated that our model could make smoother boundary prediction and achieve better clinical evaluation performance. Moreover, our methods revealed stable performance on all three tasks which is competitive with the 2D and 3D aggregation methods[5].

## 4 Conclusion

In this research, we proposed a variant of U-net with dilated pyramid pooling to strengthen the segmentation network, which achieved a fine-grained segmentation performance. Based on the Post 2017 MICCAI-ACDC challenge phase online evaluation, our method obtained results could compete with state-of-the-art methods on both clinical and geometrical evaluation metrics. Thus, we could conclude that our method has the advantage on fine-grained segmentation and might be transferred to other scenarios.

## Acknowledgement

Thanks for the support of NSFC grant 31870942.